\documentclass[letterpaper, 10 pt, conference]{ieeeconf}  

\IEEEoverridecommandlockouts                              

\usepackage{graphicx} 
\usepackage{amsmath} 
\usepackage{amssymb}  
\usepackage{multirow}
\usepackage{color}
\usepackage{subfig}
\usepackage{balance}

\title{\LARGE \bf
Multimodal Fusion with Deep Neural Networks for Audio-Video Emotion Recognition}

\author{Juan D. S. Ortega, Mohammed Senoussaoui, Eric Granger, Marco Pedersoli\\
Patrick Cardinal and Alessandro L. Koerich
\thanks{*This work was not supported by any organization}
\thanks{Juan D. S. Ortega, Mohammed Senoussaoui, Eric Granger, Marco Pedersoli, Patrick Cardinal and Alessandro L. Koerich are with \'{E}cole de Technologie Sup\'{e}rieure, University of Qu\'{e}bec, H3C 1K9, Montr\'{e}al, QC, Canada
        {\tt\scriptsize juan-david.silva-ortega.1@ens.etsmtl.ca, mohammed.senoussaoui.1@ens.etsmtl.ca, eric.granger@ etsmtl.ca, marco.pedersoli@etsmtl.ca,
        patrick.cardinal@ etsmtl.ca, alessandro.koerich@etsmtl.ca}}%
}

\begin{document}

\maketitle

\thispagestyle{empty}

\pagestyle{empty}

\begin{abstract}
This paper presents a novel deep neural network (DNN) for multimodal fusion of audio, video and text modalities for emotion recognition. The proposed DNN architecture has independent and shared layers which aim to learn the representation for each modality, as well as the best combined representation to achieve the best prediction. Experimental results on the AVEC Sentiment Analysis in the Wild dataset indicate that the proposed DNN can achieve a higher level of Concordance Correlation Coefficient (CCC) than other state-of-the-art systems that perform early fusion of modalities at feature-level (i.e., concatenation) and late fusion at score-level (i.e., weighted average) fusion. The proposed DNN has achieved CCCs of 0.606, 0.534, and 0.170 on the development partition of the dataset for predicting arousal, valence and liking, respectively.  
\end{abstract}

\section{INTRODUCTION}
State-of-the-art sensors to capture audio and video signals are paving the way for many innovative technologies that allow for unobtrusive contact-free monitoring and diagnosis. For instance, they can allow for continuous health monitoring of individuals, which is becoming increasingly important for the treatment and management of a wide range of chronic illnesses, neurological disorders, and mental health issues, such as diabetes, hypertension, asthma, autism spectrum disorder, fatigue, depression, drug addiction, etc.

It is widely believed that human environments of the future (homes, workplaces, public transit, etc.) will incorporate arrays of intelligent sensors which can support and anticipate required actions to optimally self-regulate psychological states in a pervasive and unobtrusive way. Techniques for recognizing the affective state of individuals, especially techniques for image/video and speech processing and for machine learning techniques are expected play a key role in this vision of the future.  

However, despite the sophisticated sensors and techniques, some key challenges remain for accurate affect recognition in real-world scenarios. First, it is difficult to robustly capture of spatio-temporal information, and extract common temporal features for each expression among a population that can be effectively encoded while suppressing subject-specific (intra-class) variations. There are significant spatio-temporal variations over time in facial and speech expressions according to the specific individual behaviors and capture conditions. Moreover, it is costly to create large representative audio-video datasets with reliable expert annotation as needed to design recognition models, and accurately detect of levels of arousal and valence.

In this paper, dynamic expression recognition techniques are considered to accurately asses the emotional state of subjects over time using multiple diverse modalities. A deep neural network (DNN) architecture is proposed for multimodal fusion of information extracted from voice, face and text sources for audio-video emotion recognition. Deep learning architectures have been shown to be efficient in different speech and video processing tasks \cite{Ali16,Che15,Hin12,Hua15,Taj16,silva2019emotion}. For accurate recognition, the DNN proposed in this paper provides an intermediate level of fusion, where features, classifiers and fusion function are globally optimized, in an end-to-end fashion. The proposed approach has been validated and compared to state-of-the-art systems that perform early feature-level (i.e., concatenation) and late score-level (i.e., weighted average) fusion on the AVEC Sentiment Analysis in the Wild (SEWA) database. 
 
The paper is organized as follows. Section II describes the emotion recognition techniques proposed in literature for different modalities. Section III describes the proposed DNN architecture for multi-modal fusion. Section IV presents the experimental protocol while Section V present the experiments carried out on the RECOLA dataset and the results achieved by the proposed approach for each modality (audio, video and text) as well as with their fusion. Finally, the conclusions and perspectives of future work are stated in the last section.

\section{Related Work}
Over the past decade, facial expression recognition (ER) has been a topic of significant interest. Many ER techniques have been proposed to automatically detect the seven universally recognizable types of emotions -- joy, surprise, anger, fear, disgust, sadness and neutral -- from a single still facial image \cite{cossetin2016facial,Kum15,oliveira20112d,tannugi2019memory,zavaschi2013fusion,zavaschi2011facial}. These static techniques for facial ER tend to follow either appearance-based or geometric-based approaches. More recently, dynamic techniques for facial ER has emerged as a promising approach to improve performance, where the expression type is estimate from a sequence of images or video frames captured during physical facial expression process of a subject \cite{Pan06}. This allows to extract not only facial appearance information in the spatial domain, but also its evolution in the temporal domain. Techniques are either shape-based, appearance-based or motion-based methods \cite{Guo16}. 

Shape based methods like the constrained local model (CLM) describe facial component shapes based on salient anchor points. The movement of those landmarks provides discriminant information to guide the recognition process. Appearance based methods like LBP-TOP extract image intensity or other texture features from facial images to characterize facial expressions. Finally, motion based methods like free-form deformation model spatial-temporal evolution of facial expressions, and require reliability face alignment methods. For instance, Guo \textit{et al.}~\cite{Guo16} used an atlas construction and sparse representation to extract spatial and temporal information from a dynamic expression. Although the computational complexity is higher, including temporal information along with spatial information, greater recognition accuracy was achieved compared to that of static image FER.

Automatic detection of a speaker's emotional state has gained popularity over the past few years. Different approaches have been explored to improve accuracy of depression detection and emotion detection system. Both applications have some similarities. From the depression detection side, France~\textit{et al.}~\cite{Fra00} has shown that changes in format frequencies are good indicators of depression and the propensity to commit suicide. Cummings \textit{et al.}~\cite{Cum13} reported an accuracy of approximately 70\% in a binary classification of depression (depressed and non-depressed) by using energy and spectral features. Moore \textit{et al.}~\cite{Moo03} achieved an accuracy of 75\% using statistical analysis (mean, median and standard deviation) on prosodic features such as pitch, energy or speaking rate. As is the case of almost every field in which machine learning techniques are involved, the use of neural networks has become very popular in emotion detection. Researchers have used DNNs \cite{Cardinal2015}, Long-Short Term Memory Neural Setworks (LSTMs) \cite{Hua15,Che15}, and Convolutional Neural Networks (CNNs) \cite{Zha16}.

Evidence from several studies, for either emotion or depression detection, including results from previous AVEC competitions, suggests that the accuracy and reliability of a recognition system can be improved by integrating the evidence obtained from multiple different sources of information, mainly at feature, score or decisions levels \cite{Cardinal2015,Hua15,Che15}. Consequently, there has been some recent interest in detection the emotional state through multimodal fusion, and in particular with speech and facial modalities. Kachele \textit{et al.}~\cite{Kac14} propose a hierarchical multi-classifier framework that adaptively combines input modalities by incorporating variable degrees of certainty. Vocal prosody and facial action units have also been combined to detect depression \cite{Coh09}. Meng \textit{et al.}~\cite{Men13} propose a multi-layer system using Motion History Histogram dynamical features to recognize depression from both audio and video. Nasir \textit{et al.}~\cite{Nas16} proposed a multi-modal features that capture depression behavior cues in a multi-resolution modeling and fusion framework. In particular, they employ Teager energy-based and i-vector features along with phoneme rate and duration to predict in audio, and polynomial parameterization of temporal variation and area features obtained from facial landmarks in video. Finally, Williamson \textit{et al.}~\cite{Wil14} proposed a promising system that exploits complementary multi-modal information on speech, prosody and facial action unit features to predict depression severity, although the contribution of each modality is not discussed.

Despite the sophisticated sensors and powerful techniques, some key challenges remain for the development of accurate models for emotion recognition in real-world ('in-the-wild') scenarios. During design, there is a limited amount of representative data. It is assumed that a recognition model is designed using a limited number of labeled reference samples extracted from subjects under specific conditions. Although many audio-video signals can be captured to design recognition models, they require costly expert annotation to create large-scale data sets for the detection of levels of arousal and valence. During operations, faces and speech recordings are captured using standard web cams and microphones in an operational domain (i.e., subject's office, home, etc.), under various conditions. There are significant dynamic variations over time in facial and voice expressions according to the specific capture conditions and individual behaviors. Therefore, recognition models are not representative of the intra-class variations of modalities in the operation domain. Indeed, any distributional change (either domain shift or concept drift) can degrade system performance. Recognition models require calibration and adaptation to the specific person, the sensors and computing device, and the operational environment. 

In is well known that combining spatio-temporal information from diverse modalities over time can improve the robustness and recognition accuracy. Modalities may also be dynamically combined according to contextual or semantic information, for example noise in the recording environment. For example, output responses of deep learning architectures can be combined at different layers (resolution) based on the capture conditions. 

This paper is focused on exploiting deep learning architectures to produce accurate mixtures of affect recognition systems. For instance, Kim \textit{et al.}~\cite{Kim2015} proposed a hierarchical 3-level CNN architecture to combine multi-modal sources. DNNs are considered to learn a transformation sequence with the specific objective to obtain features that will combined in one system. Since feature-level and score-level fusion do not necessarily procure a high level of accuracy, a hybrid approach is proposed where features and classifier are learned such that they are optimized for multi-modal fusion. 

\section{Multimodal Fusion with Deep NNs}

\begin{figure*}
  \centering \includegraphics[scale=0.55]{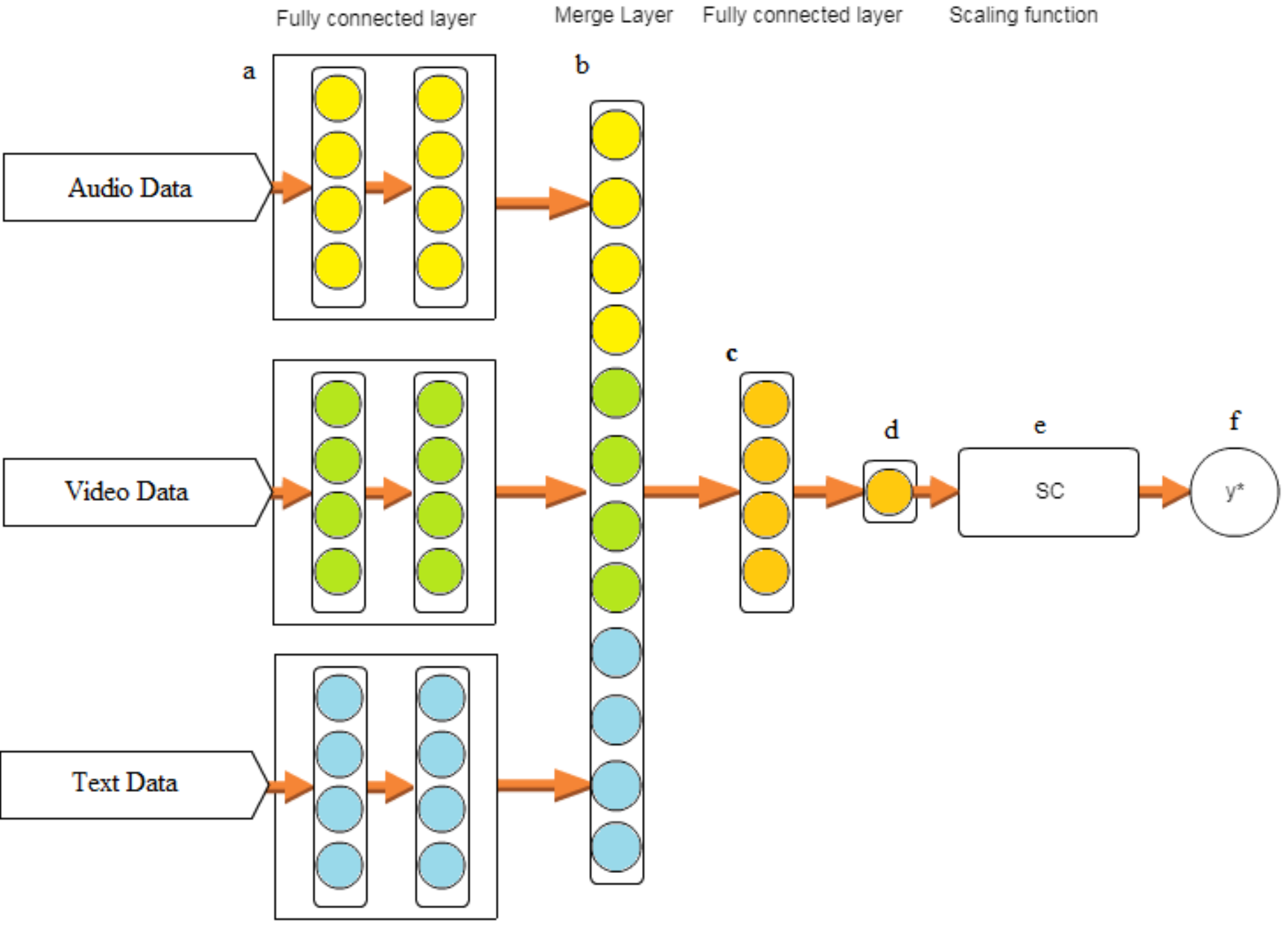}
  \caption{Proposed DNN architecture for multi modal fusion: (a) independent layers; (b) merge layer; (c) fully-connected layer; (d) regression layer; (e) scaling module; and (f) final prediction.}
  \label{fig:DNNarchit}
\end{figure*}

This paper presents a new cost-effective DNN architecture that can learn a robust mapping between a subject's spontaneous and natural behaviors from multiple sources of information and his emotional state. Given the subtle patterns of change over time in a subject's face and speech modalities (and textual information) from audiovisual recordings in the AVEC SEWA database, the system proposed in this paper learns the feature representation, and the classification and fusion functions to accurately predict the subject's levels of arousal, valence and liking. 
 
The main intuition of the proposed approach is to co-jointly learn each a discriminant representation, along with their classification and fusion functions. Each feature subset is first independently processed by one or more hidden layers. This part of the network learns the best feature representation for a given task. Then, the last hidden layers of each block are inter-connected to one or more fully connected layers that will perform the classification and fusion of representations. Globally, this network should learn how the input features can be transformed with the intention of being classified and combined, and producing one global decision. Training a hybrid classifier to combine these representations could improve the overall accuracy of the recognition system. The proposed architecture exploits three different sources of information -- audio (voice), video (face) and text. For more details on the feature sets, please refer to \cite{avec17}.

\subsubsection{Audio Features} Acoustic features consist of 23 acoustic low-level descriptors (LLDs) such as energy, spectral and cepstral, pitch, voice quality, and micro-prosodic features, extracted every 10ms over a short-term frame. Segment-level acoustic features are computed over segments of 6 seconds using a codebook of 1,000 audio words and a histogram of audio words is created. Therefore, the resulting feature vector has 1,000 dimensions. 
\subsubsection{Video Features} The video features are extracted for each video frame (frame step 20ms) and consist of three feature types: normalized face orientation in degrees; pixel coordinates for 10 eye points; pixel coordinates for 49 facial landmarks. Bag-of-video-words with separate codebooks and histograms were created for each three facial feature type, with a codebook size of 1,000 each, resulting in a 3,000-dimensional segment-level feature vector.
\subsubsection{Text Features} Text features consist of a bag-of-words feature representation based on the transcription of the speech. The dictionary contains 521 words, where only unigrams are considered. The histograms created over a segment of 6 s in time. In total, the bag-of-text-words (BoTW) features contain 521 features. 

\subsection{DNN Architecture}
The proposed DNN architecture for multi-modal fusion is shown in Fig.~\ref{fig:DNNarchit}. The DNN processes each source of information -- audio, video, and text -- separately using a pair of fully connected layers per modality (a) in order to generate and explore correlations between features of the same type. Then, a second stage (b) merges the outputs of those independent layers using a concatenation function. This layer receives the outputs of the previous step in one block, and feeds a fully connected layer (c) that correlates the essence of the modalities. The DNN output is generated by a single linear neuron (d) used as the regression value of the whole network. Finally, a scaling module (e) is used to reduce the gap between magnitudes of the predictions and the labels. Different scaling functions have been considered to yield the best results -- decimal scaling, min-max normalization and standard deviation scaling. The final prediction (f) is produced via a linear activation function (weighted sum of previous layer).  

For the training process the Mean Squared Error (MSE) was used as loss function:

\begin{equation} 
\label{equ:MSE}
MSE = \frac{1}{m} \sum_{i=1}^m (\hat{y_i} - y_i)^2
\end{equation}

\noindent where $\hat{y}$, $y$, and $m$ denote the expected values, the observed values and the number of predictions respectively. The `checkpoint' functionality (given by Keras API \cite{chollet2015keras}) was used in order to get the best model during the training process. Several training sessions were run for each dimension (arousal, valence, and liking) to ensure a stable configuration of the networks.

\section{Experimental Methodology}
\label{sec:protocol}
The experimental protocol used with RECOLA dataset involves a training set and a development set. For experiments conducted in this research, the development set has been split into two subsets. The first one contains five randomly chosen subjects out of the 14 subjects in the development set. The nine remaining subjects have been used as a test subset to evaluate the performance of the proposed approach.

For the SVR experimentation, the original protocol has been used for building unimodal and the early-fusion systems. The late-fusion approach used the first development subset to optimize each unimodal system. The fusion function has been optimized using the second development subset.

In the case of the DNN, the smaller development subset has been used to determine the optimal number of layers and neurons for each modalities as well as the number of neurons in the fusion layer. The second development subset has been used for testing the model. Note that each emotional dimension has his own architecture, described in Table~\ref{tab:dnnArch}.

\begin{table}
\renewcommand{\arraystretch}{1.2}
\caption{Description of the DNN architecture for each emotional dimension (arousal,valence and liking).}
\label{tab:dnnArch}
\begin{center}
\begin{tabular}{|l|c|c|c|c|c|c|}
\hline
Modality  		& \multicolumn{2}{|c|}{Arousal} & \multicolumn{2}{|c|}{Valence}& \multicolumn{2}{|c|}{Liking}\\
\cline{2-7}
                & L1 & L2 & L1 & L2 & L1 & L2  \\
\hline
Audio			& 50	& 50  & 200 & 200 & 50 & 50  \\
Video           & 100  & 100  & 200 & 200 & 100 & 100\\
Text            & 200  & 200  & 200 & 200 & 100 & 100 \\
\hline
Fusion Layer    & \multicolumn{2}{|c|}{100} & \multicolumn{2}{|c|}{200} & \multicolumn{2}{|c|}{50}\\ 
\hline
\end{tabular}
\end{center}
\end{table}

The evaluation measure used in the challenge is the CCC, which is defined as:

\[ \rho_{y^\prime y} = \frac{2*s_{y^\prime y}}{s^2_{y^\prime} + s^2_y+(\bar{y^\prime}-\bar{y})^2} \]

\noindent where $y^\prime$ and $y$ are the sets for which the correlation is calculated, $s^2_{y^\prime}$ and $s^2_y$ are the variances calculated on sets $y^\prime$ and $y$ respectively and $\bar{y^\prime}$ and $\bar{y}$ are the means of sets $y^\prime$ and $y$, respectively.

\subsection{Preprocessing}
\begin{figure}[htpb!]
  \centering \includegraphics[scale=0.70]{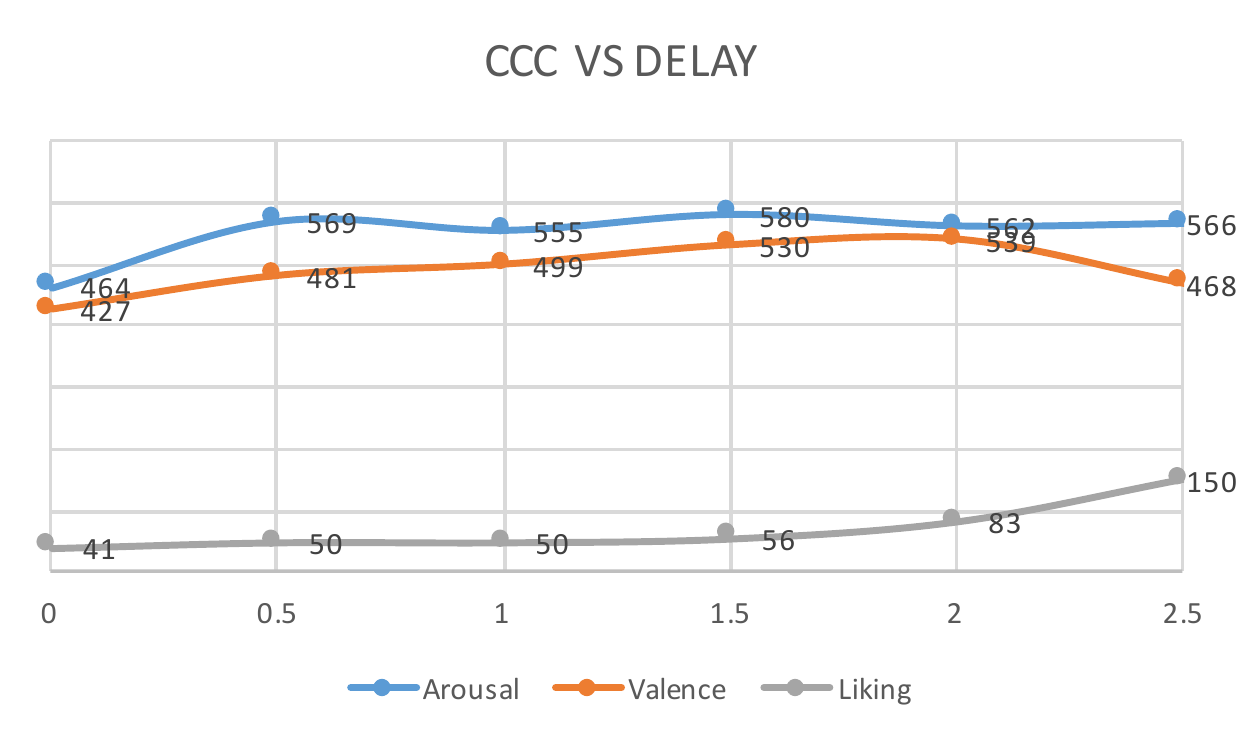}
  \caption{Effect of the delay on the CCC}
  \label{fig:cccDelay}
\end{figure}

A significant increase over the results of the CCC score appears when a delay compensation function is applied. Fig.~\ref{fig:cccDelay} shows how the CCC change in function of the delay. The suitable value for the delay compensation (optimized on the development partition) was chosen as $d=1.5$ for arousal and valance, the curves behavior shows that this point is useful in order to increase the CCC for those dimensions on the other hand the Liking predictions shows a better behavior when the delay comes to be $d=2.5$

\subsection{Postprocessing}
The output of the DNN regression layer can be any real value. However, the regression process tends to average the values obtained from the feature vectors in different ways. That process implicitly leads to an attenuation of the predicted values.  This issue can be circumvented by scaling the outputs. 

\ \\
\noindent \textbf{Min-Max Scaling}
Min-Max scaling provides a linear transformation on the original range of predictions in a pre-defined boundary. A modified value $\vec{y}_{norm}$ is obtained using Equation \ref{eq:minMax_ratio}:

\begin{equation} 
\label{eq:minMax_ratio}
\vec{y}_{norm} = \frac{\max_{l}-\min{l} \cdot \vec{y}-\min_{p}}{\max_{p}-\min_{p}}+\min_{l} 
\end{equation}

\noindent where $\max_{l}$ and $\min_{l}$ respectively are the maximum and minimum values in the training labels, $\max_{p}$ and $\min_{p}$ respectively are the maximum and minimum values of $\vec{y}$, the original prediction vector. 

\ \\
\noindent \textbf{Standard-Deviation Ratio}
This scaling method has been used in \cite{Trigeorgis2016}, with interesting results. A modified value $\vec{y}_{norm}$ is obtained using Equation \ref{eq:std_ratio}:

\begin{equation} 
\label{eq:std_ratio}
\vec{y}_{norm} = \frac{\sigma_{p}}{\sigma_{l}}\otimes \vec{y}
\end{equation}
\noindent where $\sigma_{p}$ is the standard deviation of the predictions, $\sigma{l}$ is the standard deviation of the golden standard, $\otimes$ is the element-wise multiplication operation and $\vec{y}$ is the prediction vector to be scaled. 

\ \\
\noindent \textbf{Decimal Scaling}
This scaling method moves the decimal point of values of the prediction vector $\vec{y}$. The number of decimal points moved depends on the maximum absolute value in $\vec{y}$. A modified value $\vec{y}_{norm}$ is obtained using Equation \ref{eq:dec}:
\begin{equation}
\label{eq:dec}
\vec{y}_{norm} = \frac{\vec{y}}{10^{min_{p}}}
\end{equation}

\noindent where $\vec{y}$ is the original prediction vector and $min_p$ is the smallest predicted number such that $max|\vec{y}_{norm}|<1$ and the division is performed element-wise.

\begin{table*}[htpb!]
\renewcommand{\arraystretch}{1.2}
\caption{SVR results for emotion recognition on the Development (D) partition. Performance is measured in terms of the Concordance Correlation Coefficient.}
\label{tab:result_SVR}
\begin{center}
\begin{tabular}{|l|c|c|c|c|c|c|c|c|c|c|}
\hline
Modality  		& \multicolumn{3}{|c|}{Arousal} & \multicolumn{3}{|c|}{Valence} & \multicolumn{3}{|c|}{Liking}\\
\cline{2-10}
                & No Scaling & Std Ratio & Min-Max & No Scaling & Std Ratio & Min-Max  & No Scaling & Std Ratio & Min-Max \\
\hline
D-Audio			& .361	& .400	& \bf{.449} & .412 & .418 & \bf{.420} & .037 & .028 & \bf{.040} \\
D-Video           & .455  & \bf{.464}  & .337 & \bf{.379} & \bf{.379} & .344 & \bf{.174} & .166 & .133 \\
D-Text            & .366  & \bf{.409}  & .373 & .380 & \bf{.402} & .399 & \bf{.315} & .301 & .327 \\
\hline
D-Early Fusion	& .525	& \bf{.572}	& .393 & .508 & \bf{.532} & .491 & .154 & \bf{.157} & .099 \\
D-Late Fusion     & \bf{.387}  &  .358   & .259     & \bf{.319} &  .314   & .398  & .220 &  .247    & \bf{.290} \\ 
\hline
\end{tabular}
\end{center}
\end{table*}

\begin{table*}[htpb!]
\renewcommand{\arraystretch}{1.2}
\caption{DNN results for emotion recognition on the Development (D) and Test (T) partitions. Performance is measured in terms of the Concordance Correlation Coefficient.}
\label{tab:results_DNN}
\begin{center}
\begin{tabular}{|l|c|c|c|c|c|c|c|c|c|c|}
\hline
Modality  		& \multicolumn{3}{|c|}{Arousal} & \multicolumn{3}{|c|}{Valence} & \multicolumn{3}{|c|}{Liking}\\
\cline{2-10}
                & No Scaling & Decimal  & Std Ratio & No Scaling & Decimal  & Std Ratio  & No Scaling & Decimal & Std Ratio \\
\hline
D-Early Fusion	& .542	& .542	& \bf{.565} & .467 & .492 & \bf{.500} & .145 & \bf{.198} & .185 \\
D-Proposed Fusion     & .580  &  \bf{.606}   & \bf{.606} & .530 &  .522   & \bf{.534}  & .150 &  .165    & \bf{.170} \\ 
\hline
T-Proposed Fusion  & NA & NA & .192 & NA & NA & .320 & NA & NA & .152 \\
\hline
\multicolumn{10}{l}{\scriptsize NA: Not Available}\\
\end{tabular}
\end{center}
\end{table*}

\section{Experimental Results and Discussion}

\subsection{Results with SVR}

Some preliminary experiments have been carried out using a SVR in order to provide a advanced baseline. The complexity (as provided in the baseline script), the epsilon (in the range [0.0-0.0001]) and the delay (in the range [0-3] seconds) have been optimized on the development set. Table~\ref{tab:result_SVR} shows results on the experiments conducted with the SVR framework. Note that the protocol used in the baseline paper~\cite{avec17} has been used excepting for the early fusion scheme. The first three lines present the CCC obtained on each individual modality. A first observation is that predicting the liking is a quite hard task, particularly from the audio and video. However, the text seems a much better modality for predicting liking. Moreover, the text also performs well for both the arousal and valence. A major drawback of this modality is the need of transcriptions, for which the production is both time consuming and expensive and not realistic in real-life applications. However, with the accuracy obtained with state-of-the-art speech recognition systems, the use of the text modality is not out of reach. It would be interesting to study how the errors made by a speech recognition system will affect the precision of an emotion identification system. Another observation is that the use of early fusion led to a 23\% improvement for arousal and 32\% for valence. The early fusion scheme is very harmful when applied to the liking dimension. This is mainly due to the poor performance of audio and video features for this dimension.     

\begin{figure}[htpb!]
  \centering \includegraphics[scale=0.65]{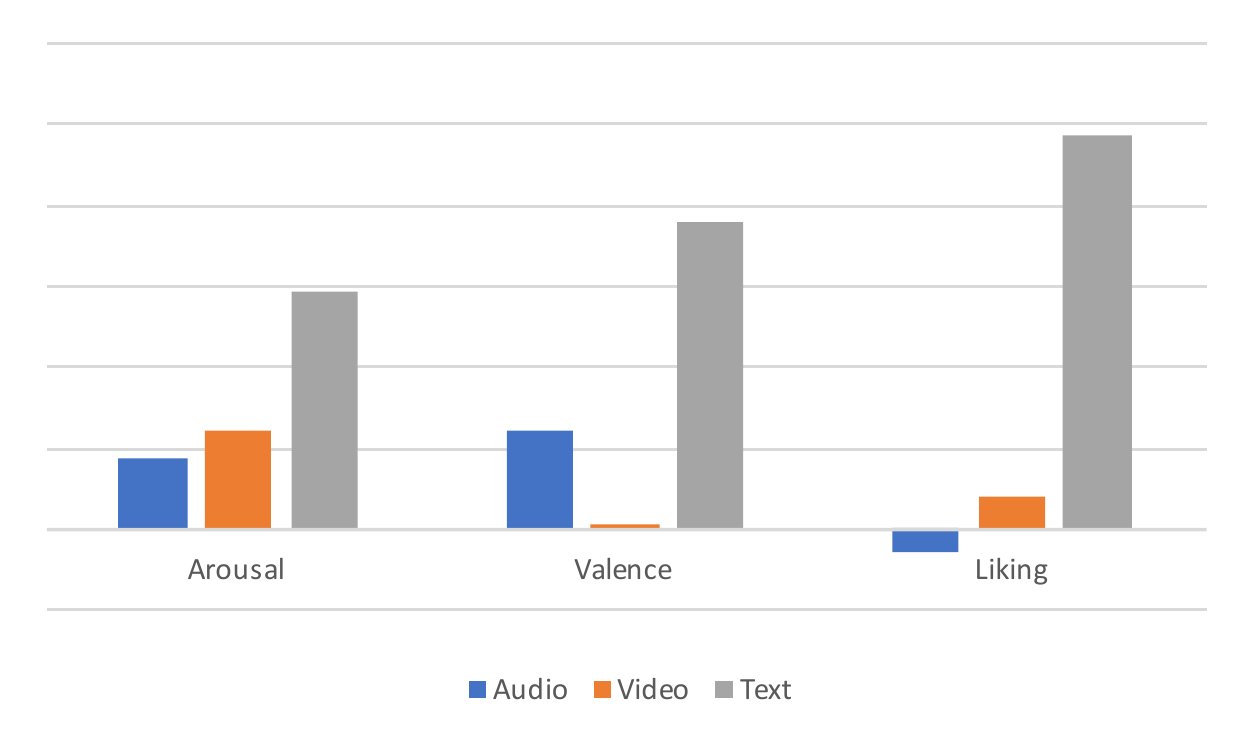}
  \caption{Relative importance of each modality in a late fusion scheme}
  \label{fig:modImp}
\end{figure}

The last experiment conducted with SVR is on the late fusion. This part has been trained using the experimental protocol explained in section~\ref{sec:protocol}. A simple linear regression\footnote{The sklearn package in python has been used to calculate the linear regression} has been used for fusion. The results is not as good as with the early fusion scheme, excepting for the liking dimension. The most interesting point with this experiment is the relative contribution of each modality in the prediction of the emotion, as shown in Fig.~\ref{fig:modImp}. Values are derived from the linear regression coefficients obtained from the multimodal late fusion model. As previously mentioned, it is clear from Fig.~\ref{fig:modImp} that text is the most important modality for all dimensions. It is particularly true for the liking, for which the text contribution is more than 97\% of the predicted value. Note that for arousal and valence, the text contributes to more than 50\% of the prediction value, which shows the importance of this modality. Another interesting point in this figure is the negative contribution of the audio in prediction of the liking, suggesting that audio (or at least the feature used to represent audio recordings) is not suitable for the prediction of the liking.

\subsection{Results with DNNs}

The first conducted experiment with a DNN was the early fusion scheme. Table~\ref{tab:results_DNN} shows the results obtained on the development set. The performance of the early fusion scheme depends on the modality. For the arousal, using whether a DNN or a SVR leads to same result. However, the SVR outperforms the DNN for the valence prediction by 6\%. On other hand, the DNN is less sensitive to unusable modality since it outperforms the SVR by 26\%.  

The results achieved with the proposed approach are very interesting since they outperform both the DNN and the SVR for the arousal and valence dimensions. However, late fusion is still the better choice for liking prediction. This means that, even with the proposed architecture, the DNN is not able to ignore modalities that are useless or hurtful. However, this architecture works better than early fusion with DNN. The performance on the test set are a bit low. The best explanation for this results is over-fitting. However, since the results on the development set suggest that the proposed approach provide a relatively high level of performance, more experiments will be conducted to understand this problem. 

\section{Conclusion and Future Work}
In this paper, a new DNN architecture is proposed for the estimation of the emotional state of a subject. It integrates three different modalities: vocal signal, facial features and textual transcript of the vocal signal. Each modality is first encoded independently with two fully connected layers and then merged into a single representation which is then used for estimating the emotional state of a subject. The network, trained in a end-to-end fashion, provides higher CCC than other proposed architectures. Further improvement are expected by using a proper normalization of the input features as well as by adding a temporal smoothing of the regressed output.
For future work we plan to extend our work to include a final stage based on a recurrent neural network that can learn temporal patterns of emotions and therefore improve the overall system accuracy. Also, we will evaluate the performance of visual features extracted from a convolutional NN explicitly trained for the task. 
 
\balance



\bibliographystyle{ieee}


\end{document}